\documentclass{article} 
\usepackage[nonatbib,preprint]{neurips_2019}

\usepackage{amsmath,amsfonts,bm}

\def\eqref#1{equation~\ref{#1}}

\def\1{\bm{1}}

\DeclareMathAlphabet{\mathsfit}{\encodingdefault}{\sfdefault}{m}{sl}
\SetMathAlphabet{\mathsfit}{bold}{\encodingdefault}{\sfdefault}{bx}{n}

\usepackage{hyperref}
\usepackage{url}

\usepackage{epsfig}
\usepackage{graphicx}
\usepackage{amsmath}
\usepackage{amssymb}
\usepackage{amsmath}
\usepackage{tabularx}
\usepackage{float}
\usepackage[caption = false]{subfig}
\usepackage{adjustbox}
\usepackage[none]{hyphenat}

\title{Lift-the-flap: what, where and when for \\ context reasoning}

\author{Mengmi Zhang$^{1}$  \quad Claire Tseng$^2$  \quad Karla Montejo$^3$  \quad Joseph Kwon$^4$   \quad Gabriel Kreiman$^1$ \\ \\
$^1$ Boston Children's Hospital, Harvard Medical School \\
$^2$ Harvard College \\
$^3$ Mayo Clinic Graduate School of Biomedical Sciences \\
$^4$ Yale University \\
Address correspondence to \texttt{gabriel.kreiman@tch.harvard.edu}
}

\begin{document}

\maketitle

\begin{abstract}

Context reasoning is critical in a wide variety of applications where current inputs need to be interpreted in the light of previous experience and knowledge. Both spatial and temporal contextual information play a critical role in the domain of visual recognition. Here we investigate spatial constraints (what image features provide contextual information and where they are located), and temporal constraints (when different contextual cues matter) for visual recognition. The task is to reason about the scene context and infer what a target object hidden behind a flap is in a natural image. To tackle this problem, we first describe an online human psychophysics experiment recording active sampling via mouse clicks in lift-the-flap games and identify clicking patterns and features which are diagnostic for high contextual reasoning accuracy. As a proof of the usefulness of these clicking patterns and visual features, we extend a state-of-the-art recurrent model capable of attending to salient context regions, dynamically integrating useful information, making inferences, and predicting class label for the target object over multiple clicks. The proposed model achieves human-level contextual reasoning accuracy, shares human-like sampling behavior and learns interpretable features for contextual reasoning.
\end{abstract}

\section{Introduction}

The tiny object on the table is probably a spoon, not an elephant. Objects do not appear in isolation. Instead, they co-vary with other objects, their sizes and colors usually respect regularities with respect to nearby elements, and objects tend to appear at specific locations within a scene. Humans exploit these contextual associations. Contextual analyses based on the statistical summary of object  relationships, provide an  effective source of information for perceptual inference tasks, such as object detection (\cite{torralba2003contextual,park2010multiresolution, hoiem2005geometric, torralba2010using,liu2018structure}), scene classification (\cite{gonfaus2010harmony,torralba2005contextual,yao2012describing}), semantic segmentation (\cite{yao2012describing}), and visual question answering (\cite{teney2017graph}).

\begin{figure*}[t]
 \centering
     \subfloat[Lift-the-flap game]{\includegraphics[width= 4.2cm, height = 2.8cm]{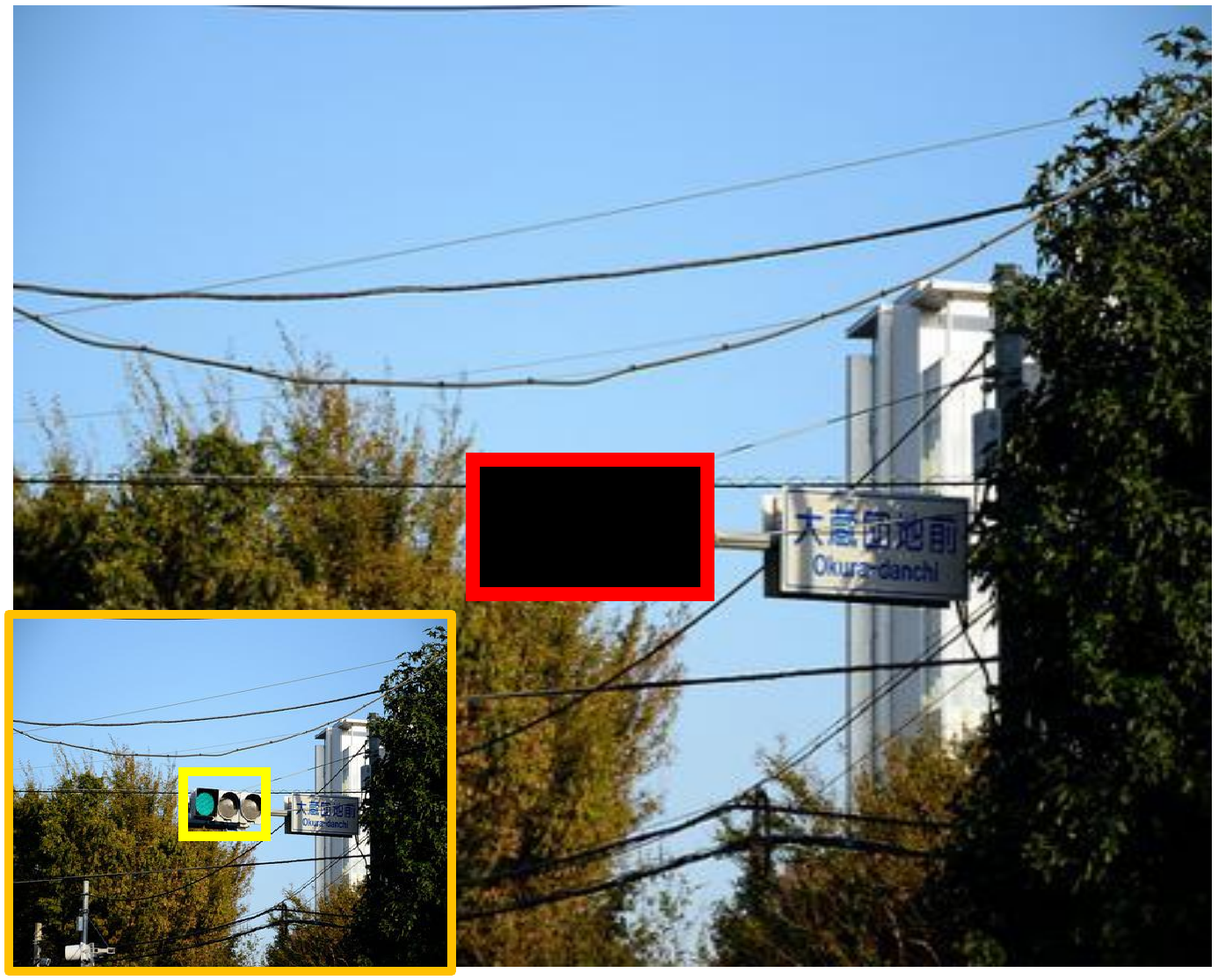}\label{fig:intro}}\hspace{0.2cm}
     \subfloat[Schematics of Behavioral Experiment]{\includegraphics[width= 7.0cm, height = 2.8cm]{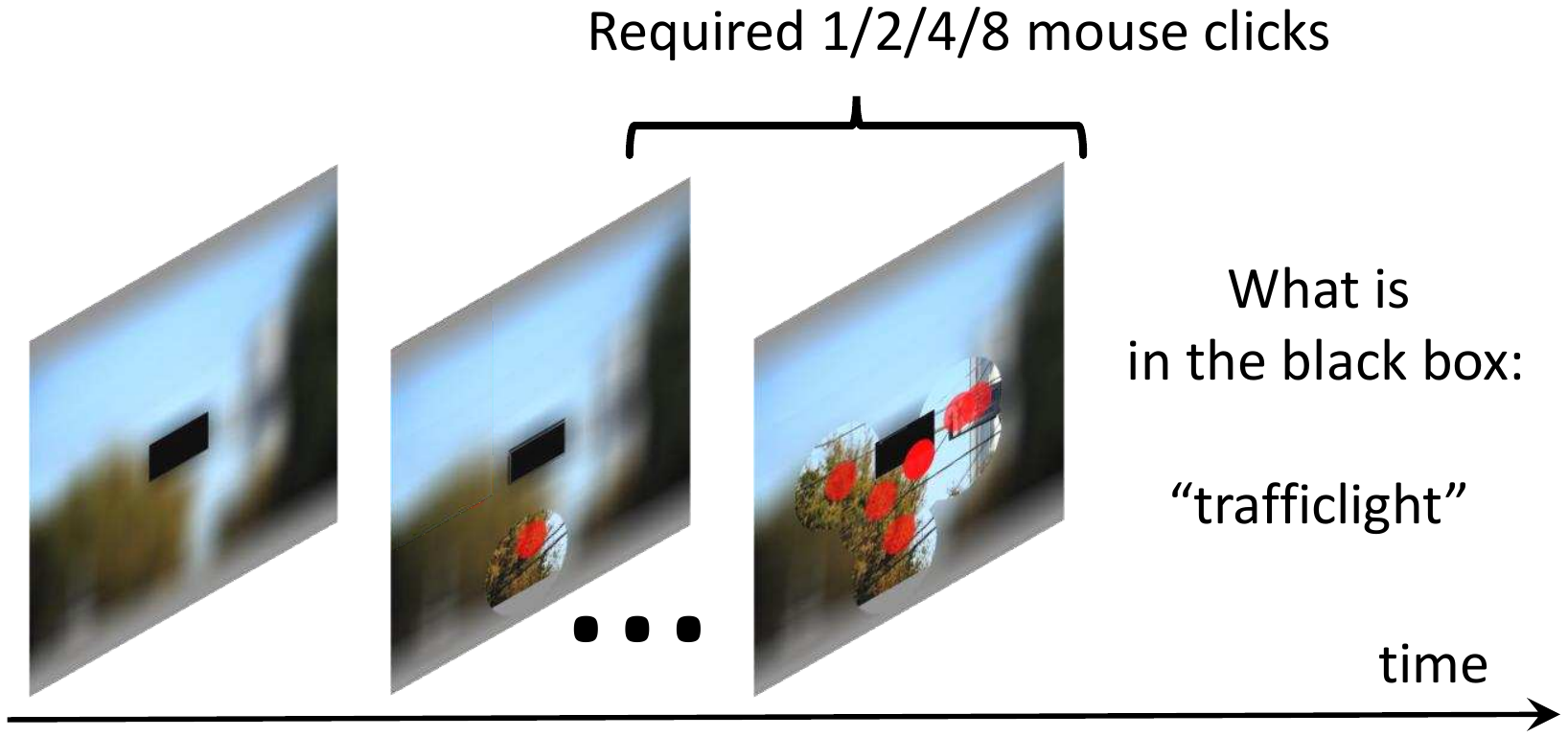}\label{fig:psycho}}
     \caption{\textbf{Schematic of the lift-the-flap task and human behavioral experiment}. (a) The task requires subjects to capitalize on the scene context in a natural image to infer what is behind the black box (the hidden target). The original image (bottom left) reveals the target object (``traffic light"); this image was \emph{not} shown in the actual experiments. (b) A blurred image with the hidden target was presented to the subject. To identify contextual areas of importance surrounding the hidden target, subjects used the computer mouse to click on image a pre-specified number of times (red dots). Upon clicking a certain location, a circle of fixed radius was revealed at high resolution. After the required number of clicks, subjects typed a noun describing the hidden target object. The experiments were conducted online using Amazon Mechanical Turk \cite{turk2012amazon} on 100 subjects, 50 trials per subject. Figure 3 also shows results for a variation of the experiment conducted in the lab while tracking eye movements.}
 \vspace{-4mm}\label{fig:introall}
 \end{figure*}

An example of how contextual information is incorporated during object recognition is lift-the-flap books, where a flap covers part of the page. Children make guesses about what is behind the flap based on the context and check their answers by lifting the flap (Figure~\ref{fig:intro}). Here we investigate \emph{what} image features matter for contextual reasoning and \emph{where} those features are with respect to the target object of interest. Furthermore,  scene interpretation in humans involves a sequence of eye movements \cite{zhang2018finding}, each one of these image samples providing additional context to inform interpretation of the contents of the next location. Therefore, we also investigate \emph{when} scene information matters for context reasoning.

To tackle the problem of contextual reasoning, we introduce the lift-the-flap task and conduct online psychophysics experiments where subjects make mouse clicks while they explore important contextual cues to identify a hidden target (Figure~\ref{fig:psycho}). We investigate the  contextual reasoning strategies observed from human active sampling patterns. As a proof of concept, we propose a recurrent attention model (ClickNet), to automatically learn these contextual reasoning strategies. The model guides attention towards regions with informative context, decides where to sample the image, and makes inferences about the target behind the flap. The learnt sampling patterns and predicted class labels by ClickNet share remarkable similarities with human behavior.

\section{Related Works}
\vspace{-2mm}
\subsection{Role of Context in Human Vision}

The cognitive science literature has shown that contextual information affects the efficiency of several visual processes \cite{auckland2007nontarget,biederman1982scene,hollingworth1998does,bar2003cortical,goh2004cortical,aminoff2006parahippocampal}, such as object recognition \cite{auckland2007nontarget}, object detection \cite{biederman1982scene,hollingworth1998does}, visual working memory \cite{friedman1979framing,aminoff2006parahippocampal} and visual search \cite{henderson1999effects}. Objects appearing in a familiar background can be detected more accurately and processed more quickly than objects appearing in an incongruent scene. Here we focus on what visual features contribute to contextual reasoning, which parts of image regions attract humans' attention for making inferences, and the dynamic sequence of directed sampling needed for making inferences about a hidden target.
\vspace{-2mm}
\subsection{Role of Context in Computer Vision}

Contextual reasoning about objects and relations is critical to machine vision. In fact, many object recognition studies using natural image datasets such as ImageNet, rely implicitly but strongly on contextual feature regularities \cite{geirhos2018imagenet,brendel2019approximating}.
Several studies employ contextual information in order to improve object detection \cite{park2010multiresolution, hoiem2005geometric, torralba2010using,liu2018structure}. The types of contexts can be exploited in the form of global scene context \cite{torralba2010using}, ground plane estimation \cite{park2010multiresolution}, geometric context \cite{hoiem2005geometric}, relative location \cite{desai2011discriminative}, 3D layout \cite{lin2013holistic}, and spatial support and geographic information \cite{divvala2009empirical}. In \cite{gonfaus2010harmony,yao2012describing,ladicky2010graph}, researchers proposed Conditional Random Field (CRF) models that reason jointly across multiple computer vision tasks in image labeling and scene classification. Additionally, \cite{mottaghi2014role} studies the role of context in both object detection and semantic segmentation tasks, demonstrating improved  performance in both tasks compared to raw image features. Recently, several neural network architectures incorporating contextual information have been successfully applied in object priming \cite{torralba2003contextual}, place and object recognition \cite{wu2018learning,torralba2005contextual}, object detection \cite{liu2018structure}, and visual question answering \cite{teney2017graph}. Here we focus on developing a biologically inspired computational model for contextual reasoning that can automatically and dynamically sample image regions of interest,  integrating information in space and time to make inferences about a hidden object. Additionally, we compare the model's performance against human behavior in the same task.



Several interesting approaches have combined graphical models with deep neural networks for structural inference, primarily in structured prediction tasks \cite{marino2016more,choi2012unified,chen2018deeplab,teney2017graph,hu2016learning,battaglia2016interaction,xu2017scene}. \cite{hu2016learning} designed a structured model to improve classification performance by leveraging relations among scenes, objects, and their attributes. A structured inference model is also used in \cite{choi2012unified,deng2016structure} to analyze relations in group activity recognition. Several works, like Structural-RNN \cite{jain2016structural} and Interaction Net \cite{battaglia2016interaction}, combine the power of spatiotemporal graphs and sequence learning, and evaluate the model from motion prediction to object interactions. These works assume full contextual information is available, while in our experiment we consider only partial contextual information that is sequentially revealed after a mouse click. \cite{chen2018deeplab} proposed DeepLab which inputs the response at the final layer of a deep neural network to a CRF model for semantic image segmentation. Subsequently, \cite{schwing2015fully,zheng2015conditional} transformed the CRF model into a Recurrent Neural Network in an end-to-end fashion. Breaking away from this previous work where graph optimization is performed globally, our proposed model selects important visual features using an attention mechanism and integrates partial information over multiple steps, which is computationally more efficient and accurate in the current task (Section~\ref{sec:results}).

\section{Lift-the-flap task}

\subsection{Human Behavioral Experiments}

Subjects were presented with a natural image where one object was hidden by a rectangular black box and everything else was blurred. They were allowed a fixed number of mouse clicks between 1 and 8, each click revealed part of the image in high resolution. After the target number of clicks, they had to provide a single word to describe the object hidden behind the black box (Figure~\ref{fig:psycho}). The clicking experiments were run on Amazon Mechanical Turk \cite{turk2012amazon}. The stimulus set consisted of 573 images from the test set of the MSCOCO Dataset \cite{lin2014microsoft}, spanning 80 object categories. This dataset has been widely used for object recognition and detection studies \cite{lin2014microsoft}. We constrained the stimulus set to have a uniform distribution of 6 - 8 target objects per category. To avoid any potential memory effects, subjects were only exposed to each image once. The trial presentation order was randomized.


\subsection{Ground Truth Responses}

In contrast to other experiments where subjects are forced to perform N-way categorization (e.g., \cite{tang2018recurrent}), here there were no constraints on what words subjects could use to describe the hidden object in the experiment. This probing mechanism was implemented for two reasons: first, it is difficult for humans to memorize 80 object classes in advance and there could be non-uniform memory effects impacting the results; second, we were concerned that presenting humans with an 80-choice question in each trial could introduce biases in their inference and decision processes.

We could not simply use the 80 category labels to evaluate performance because subjects could use other similar words or synonyms and we are interested in the context reasoning process rather than the subjects' language abilities. Therefore, to evaluate humans' performance, we separately collected a distribution of ground truth answers for each hidden target by presenting to 10 \emph {other} subjects, who did not participate in the main task, the same set of images with the target objects highlighted by a bounding box (not hidden). During the lift-the-flap task, a response was considered to be correct if it matched any of the ground truth labels, allowing for plurals and misspellings.

\subsection{Evaluation Metrics}\label{subsec:evaluationmetrics}

We introduce several evaluation metrics to measure contextual reasoning accuracy and to compare the consistency of mouse clicking patterns between humans and computational models. We evaluated ClickNet on the MSCOCO Dataset using the typical classification accuracy measure. In Fig~\ref{fig:whatcoratio}, we report \textbf{top-1 classification accuracy} as a function of \textbf{context-object ratio}. The context-object ratio is defined as the total area of the image \emph{excluding} the hidden target divided by the hidden target object size. For example, a context-object ratio of 1 implies that the size of the black box and the size of the contextual information is the same (see Figure~\ref{fig:whatcoratio} for example images with different context-object ratios).

To measure the degree of consistency between two mouse clicking patterns (human versus human; or human versus computational models), we computed the minimum Euclidean distance between the sequence of clicks in each trial, regardless of order. The smaller the median in the distribution of distances, the more similar the two mouse clicking patterns are.

\begin{figure*}[t]
 \begin{center}
 \includegraphics[width=12.5cm, height=4cm]{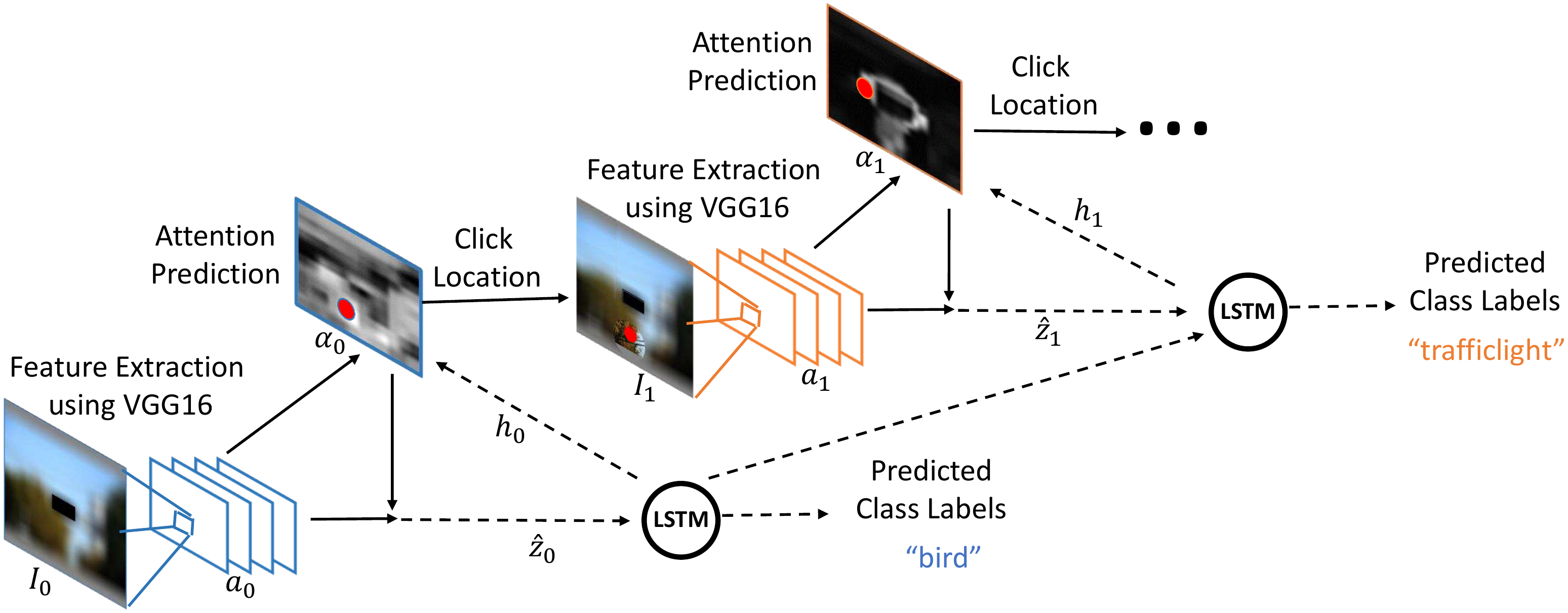}
 \end{center}
    \caption{\textbf{Architecture overview of the ClickNet model}. The diagram depicts the iterative modular steps carried out by ClickNet for contextual reasoning over multiple clicks in the lift-the-flap task (Figure~\ref{fig:psycho}). ClickNet consists of 3 main modules: feature extraction, attention, and recurrent memory. For illustrative purposes, only the first and second clicks in a trial are shown here.
    ClickNet performs feature extraction using the VGG16 network pre-trained on ImageNet and produces feature maps $a_0$. Conditioned on the hidden state $h_0$ and feature maps $a_0$,  ClickNet produces an attention map $\alpha_0$, which is used to select the next click location (red dots) and to modulate the feature maps for contextual reasoning (Figure~\ref{fig:modelattention}).
    The recurrent network in the LSTM module (Figure~\ref{fig:modellstm}) integrates over time the attentionally modulated feature maps $\mathbf{\widehat{z}_0}$, and outputs a predicted class label after each click (here ``bird" and ``trafficlight" before the 1st and 2nd clicks).
    After the first click, the input image gets updated with parts at clicked locations revealed in high resolution.  These three modular steps repeat until the specified number of clicks have been made. See supplementary figures~\ref{fig:modelattention} and~\ref{fig:modellstm} for implementation details on the attention and LSTM modules, respectively.}
 \vspace{-3mm}
 \label{fig:model}
 \end{figure*}

\section{ClickNet Architecture}\label{sec:model}

We propose a recurrent neural network for context reasoning (ClickNet), extending previous work on image captioning \cite{xu2015show}. ClickNet integrates attention-modulated context information over multiple clicks, makes a decision about the next click location based on the attention map, and infers the class label of the hidden target after every click (Figure~\ref{fig:model}).

As in the human psychophysics experiment (Figure~\ref{fig:psycho}), ClickNet is first presented with a blurred image $I_0$, which is the original image $\mathbf{I}$ with uniform gaussian blur and where the target object is covered by a black box. ClickNet  makes the first attempt to predict a class label $y_0$ out of a pre-defined set of $C$ object classes and decides its first click location $m_1$. In every trial, over a series of $T$ clicks,  the input image $I_t$ to ClickNet gets updated with circular regions of constant radius $R$ centered at all previous click locations $M=\{m_1,...m_{T}\}$, revealed in its original resolution in $\mathbf{I}$. The black box is constant and none of its content is ever revealed, even if the model opts to click within the box or if the circle centered on the click encompasses part of the box.



\subsection{Convolutional Feature Extraction}

At each time $t$ where $t \in \{0,...,T\}$, ClickNet takes $I_t$ as input and uses a feed-forward convolutional neural network to extract feature maps $a_t$. We use the  VGG16 network \cite{simonyan2014very}, pre-trained on ImageNet \cite{deng2009imagenet}. To focus on specific parts of the image and select features at those locations, we have to preserve the spatial organization of features; thus, ClickNet uses the output feature maps at the last convolution layer of VGG16.

A feature vector $\mathbf{a_{ti}}$ of dimension $D$ represents the part of the image $I_t$ at location $i$, where $i=1,..,L$ and $L= W \times H$ and $W$ and $H$ are the width and height, respectively, of the feature map:
\begin{equation}
a_t = \{ \mathbf{a_{t1}},...,\mathbf{a_{tL}} \}, \quad \mathbf{a_{ti}} \in \mathbb{R}^D
\end{equation}

\subsection{Attentional Modulation}

We use a ``soft-attention" mechanism as introduced by \cite{ba2014multiple} to compute ``the context gist" $\mathbf{\widehat{z}_t}$ on $I_t$ (Figure~\ref{fig:modelattention}). For each location $i$ in $a_t$, the attention mechanism generates a positive scalar $\alpha_{ti}$, representing the relative importance of the feature vector $\mathbf{a_{ti}}$ for context reasoning. This relative importance $\alpha_{ti}$ depends on the feature vectors $\mathbf{a_{ti}}$, combined with the hidden state at the previous step $\mathbf{h_{t-1}}$ of a recurrent network described below:
\vspace{-2mm}
\begin{equation}\label{equ:attention}
e_{ti} = A_h \mathbf{h_{t-1}} + A_a \mathbf{a_{ti}}, \quad \alpha_{ti} = \frac{\exp(e_{ti})}{\sum_{i=1}^L \exp(e_{ti})}
\end{equation}
\vspace{-2mm}

where $A_h \in \mathbb{R}^{1\times n}$ and $A_a \in \mathbb{R}^{1\times D}$ are weight matrices  initialized randomly and to be learnt. Because not all attended regions might be useful for context reasoning, the soft attention module also predicts a gating vector $\beta_t$ from the previous hidden state $h_{t-1}$, such that $\beta_t$ determines how much the current observation contributes to the context vector at each location: $\beta_{t} = \sigma(W_{\beta} \mathbf{h_{t-1}})$, where $W_{\beta} \in \mathbb{R}^{L\times n}$ is a  weight matrix and each element $\beta_{ti}$ in $\beta_t$ is a gating scalar at location $i$. As also noted by~\cite{xu2015show}, $\beta_t$ helps put more emphasis on the salient objects in the images. Once the attention map $\alpha_t$ and the gating scale $\beta_t$ are computed, the model applies the ``soft-attention" mechanism to compute  $\mathbf{\widehat{z}_t}$ by summing over all the $L$ regions in the image:
\vspace{-2mm}
\begin{equation}\label{equ:contextvec}
\mathbf{\widehat{z}_t} = \sum_{i=1}^L \beta_{ti} \alpha_{ti} \mathbf{a_{ti}}
\end{equation}
\vspace{-2mm}
The next click location $m_{t+1}$ corresponded to the maximum on the attention map:

\begin{equation}\label{equ:mouseclick}
m_{t+1} = \arg \max_i \alpha_{ti}
\end{equation}

The attentional module is smooth and differentiable and ClickNet can learn all the weight matrices in an end-to-end fashion via back-propagation.

\subsection{Recurrent Connections using long short-term memory (LSTM)}

We use a long short-term memory (LSTM) network to output a predicted class label $y_t$ based on the previous hidden state $\mathbf{h_{t-1}}$ and the context gist vector $\mathbf{\widehat{z}_t}$ for $I_t$ (Figure~\ref{fig:modellstm}). Our implementation of LSTM closely follows \cite{zaremba2014recurrent} where $T_{s,t}:\mathbb{R}^s \rightarrow \mathbb{R}^t$ defines a linear transformation with learnable parameters. The variables $\mathbf{i_t}, \mathbf{f_t}, \mathbf{c_t}, \mathbf{o_t}, \mathbf{h_t}$ represent the input, forget, memory, output and hidden state of the LSTM respectively:

\begin{equation}
\begin{pmatrix}
\mathbf{i_t} \\
\mathbf{f_t} \\
\mathbf{o_t} \\
\mathbf{g_t}
\end{pmatrix}=
\begin{pmatrix}
\sigma \\
\sigma \\
\sigma \\
\tanh
\end{pmatrix}T_{D+n,n}
\begin{pmatrix}
\mathbf{\widehat{z}_t},
\mathbf{h_{t-1}}
\end{pmatrix}
\end{equation}

\begin{equation}
\mathbf{c_t} = \mathbf{f_t} \bigodot \mathbf{c_{t-1}} + \mathbf{i_t} \bigodot \mathbf{g_t}, \quad \mathbf{h_t} = \mathbf{o_t} \bigodot \tanh(\mathbf{c_t})
\end{equation}


where $n$ is the dimensionality of LSTM, $\sigma$ is the logistic sigmoid activation, and $\bigodot$ indicates element-wise multiplication.

To cue ClickNet about the location of the hidden target, we initialize the memory state $\mathbf{c_0}$ and hidden state $\mathbf{h_0}$ of the LSTM based on a binary mask that contains zeros everywhere and ones in the hidden target location. Specifically, $\mathbf{c_0}$ and $\mathbf{h_0}$ are predicted by an average of all feature vectors $a_0$ over all $L$ locations with two separate linear transformations $W_{c0} \in R^{n\times D}$ and $W_{h0} \in R^{n\times D}$:

\begin{equation}\label{equ:hc0init}
\mathbf{c_0} = W_{c0} (\frac{1}{L} \sum_{i}^L \mathbf{a_{0i}}), \quad \mathbf{h_0} = W_{h0} (\frac{1}{L} \sum_{i}^L \mathbf{a_{0i}})
\end{equation}


To predict the class label $y_t$ of the hidden target, the LSTM computes a classification vector where each entry denotes a class probability given the hidden state:
\begin{equation}\label{equ:classpred}
y_t = \arg \max_c p(y_c), \quad p(y_c) \propto L_h \mathbf{\mathbf{h_t}}
\end{equation}
where $L_h \in \mathbb{R}^{C \times n}$ is a matrix of learnt parameters initialized randomly.

\subsection{Training and Implementation Details}

In training the model, we introduced a regularization to constrain $\sum_t \alpha_{ti} = 1$. This is to encourage ClickNet to acquire as much context information as possible by exploration. This regularization term was empirically important to improve the context reasoning accuracy. We trained ClickNet end-to-end by minimizing the cross entropy loss between the predicted label $y_t$ at each time step $t$ and the ground truth label $x$, and a regularization term for exploration:
\begin{equation}
LOSS = \sum_{t=1}^T (-\log(P(y_t|x))) + \lambda \sum_i^L(1-\sum_t^T \alpha_{ti})^2
\end{equation}

We used all images from the MSCOCO training set for training and validating all models. On every training image, each object can be blocked out as the hidden target. The input image size to ClickNet was $400 \times 400$ pixels. We used a Gaussian filter of size $51 \times 51$ with variance $64$ pixels to blur the images. The radius $R$ of the circular region revealed by each click was $55$ pixels. The hyper-parameter for the regularization term in the loss functino was $\lambda = 2$. As in the human psychophysics experiments (Fig~\ref{fig:psycho}), in each trial, we set the total number of time steps $T=8$ for training ClickNet (ClickNet predicts the label after the 1st click at $T=1$). The dimension of the LSTM module was $n = 512$. The feature maps extracted from the last convolution layer was of size $2048 \times 28 \times 28$, and the total number of locations was $L = 28 \times 28 = 784$. The Adam optimizer \cite{kingma2014adam} was used with a learning rate of $10^{-4}$ to fine-tune the VGG16 network, and a learning rate of $4\times 10^{-4}$ to train the attentional module and the LSTM module. The network was developed in Pytorch, based on \cite{xu2015show}. All source code for our proposed architecture, and the data from the psychophysics experiments will be released publicly upon publication.

\begin{figure*}[t]
\captionsetup[subfloat]{labelformat=empty,position=top}
 \centering
     \subfloat[]{\includegraphics[width= 5.5cm, height = 6.1cm]{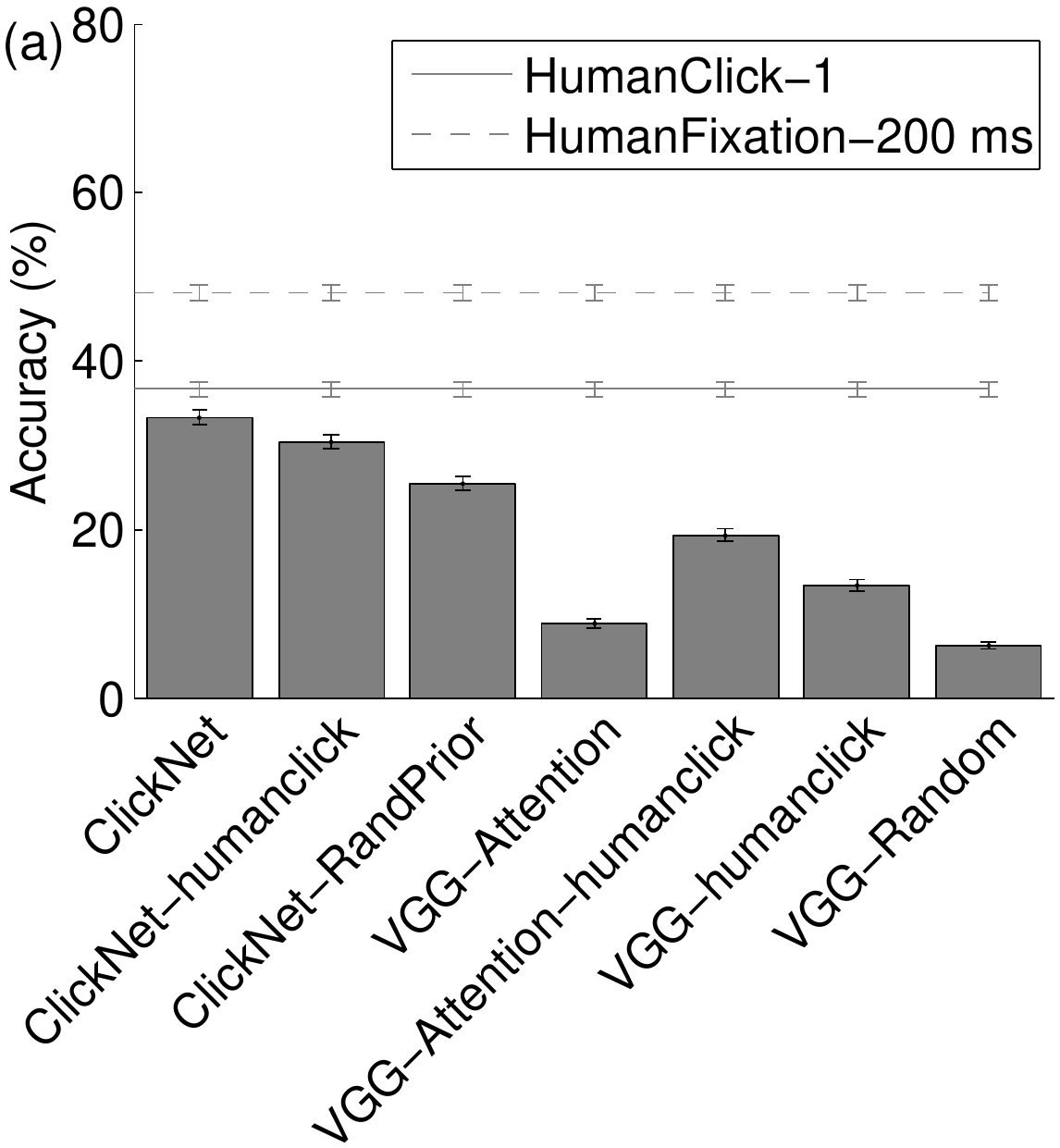}\label{fig:overallaccA}}\hspace{0.8cm}
     \subfloat[]{\includegraphics[width= 5.5cm, height = 6.1cm]{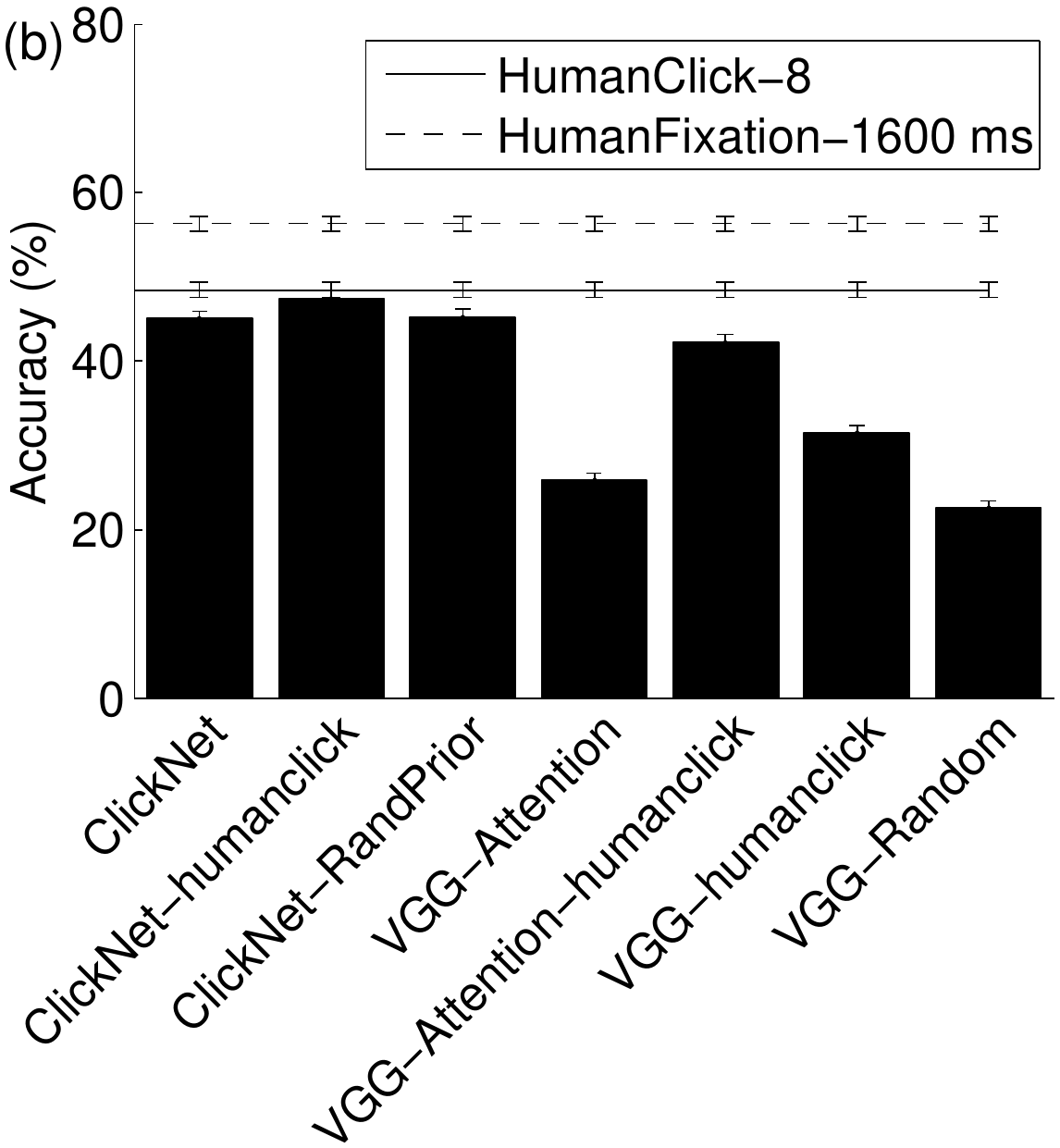}\label{fig:overallaccB}}
     \caption{\textbf{Contextual reasoning accuracy of humans and models}.     Performance for humans (horizontal lines) and models (bars) for \textbf{(a)} 1 click (gray), and \textbf{(b)} 8 clicks (black) (Fig~\ref{fig:overallperformancefull} shows results for 2 and 4 clicks, and additional comparison models). Section~\ref{subsec:evaluationmetrics} defines the evaluation metric and Section~\ref{subsec:varaitionsclicknets} describe each model. Error bars denote SEM across images.}
 \vspace{-2mm}\label{fig:overallperformance}
 \end{figure*}

 \begin{figure*}[t]
 \begin{center}
 \includegraphics[width=13.7cm, height = 3.5cm]{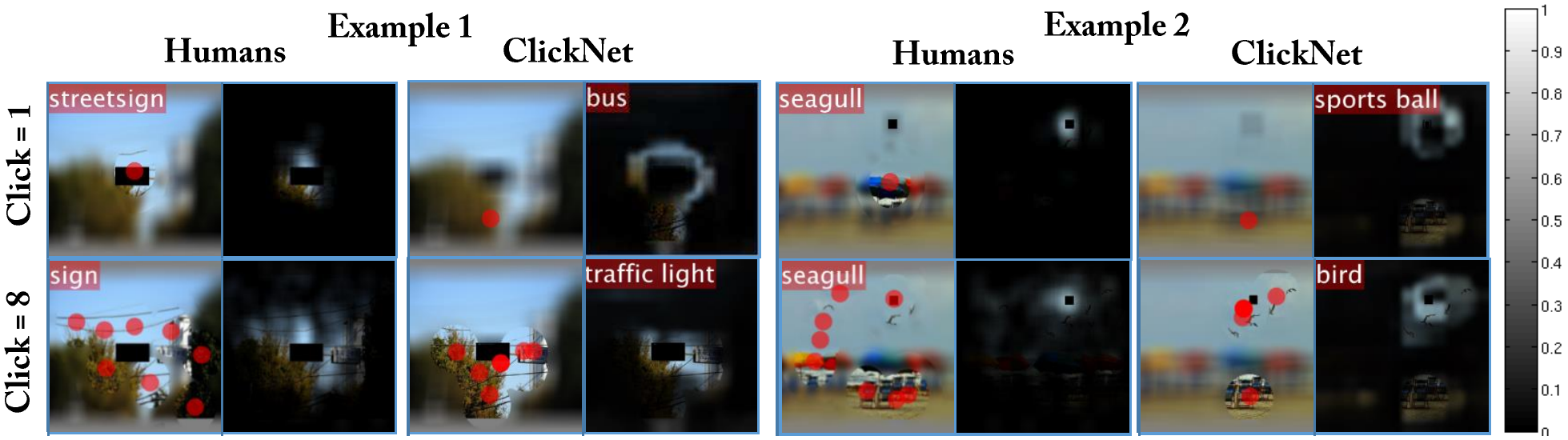}
 \end{center}
    \caption{\textbf{Example visualization for humans and ClickNet}. Two example trials (first four columns is example 1, last four columns is example 2), either with 1 click (rows 1) or 8 clicks (rows 2), for one human (columns 1, 2, 5, 6) or ClickNet (column 3, 4, 7, 8) (Fig~\ref{Sfig:examplesall} in Appendix shows results for 2 and 4 clicks). Red dots denote clicked locations. Top-left corner shows output labels after the required number of clicks. Column 2 and 6 show the  mouse click maps aggregated over  subjects. Brighter regions denote more mouse clicks (see scale bar on right). Column 4 and 8 show the attentional map predicted by ClickNet. Brighter regions denote higher attentional values.}
 \vspace{-3mm}
 \label{fig:examples}
 \end{figure*}

\subsection{Variations of Proposed Network Architecture and Comparative Methods}\label{subsec:varaitionsclicknets}

Previous work has shown that it is possible to augment vision systems with human perceptual supervision on many difficult computer vision tasks, such as \cite{vondrick2015learning,kovashka2016crowdsourcing}. One central goal in our study is to investigate what, where, and when matter for human contextual reasoning in the lift-the-flap game and whether these factors could help improve current machine learning algorithms. We now introduce two variations of ClickNet with human inputs at the \textbf{testing} stage:

\vspace{-4mm}
  \paragraph{ClickNet-humanclick.} Instead of clicking at the location with highest activation value on the attention map predicted by ClickNet, we substitute the input with human clicking images.
  \vspace{-4mm}

  \paragraph{ClickNet-RandPrior.} We observe strong spatial bias in the human clicking patterns where most of the clicks tend to be nearby the hidden target (see Sec~\ref{subsec:spatialprior} for more discussions). To test this is a useful clicking strategy, we generate random clicks along either side of the black bounding box and use these clicking images with strong spatial prior as inputs to ClickNet.

To study the role of attention and recurrent connections, we introduced two ablated models.
\vspace{-4mm}

  \paragraph{Variations of VGG16.} One intuitive way of solving the context reasoning problem is to use a feed-forward object recognition network pre-trained on ImageNet, e.g. VGG16 \cite{simonyan2014very}, and fine-tune it to classify the hidden target on MSCOCO dataset. During training, the input to the network was an image where one object on the image was randomly covered with a black bounding box.  We tested the performance of this alternative model on the 573 images selected for human psychophysics experiments with different input variations: human clicking images (\textbf{VGG-humanclick}), the blurred images (\textbf{VGG-Blur}), the full-resolution images (\textbf{VGG-Fullres}) and images with random clicks (\textbf{VGG-Random}).

  \vspace{-4mm}

  \paragraph{VGG-Attention.} Previous work has demonstrated the efficiency of attention in computer vision tasks \cite{nguyen2018attentive}, such as question answering and image captioning \cite{xu2015show}. To study the effect of attention in contextual reasoning, we added an attention module to the end of VGG16. To make the complexity of the architecture comparable with ClickNet, we added the same number of fully connected layers as in the LSTM module. As in ClickNet, we used the location with the highest activation value on the attention map to predict the next click. VGG-Attention takes the updated image as input and iteratively predicts the hidden target label. In contrast to ClickNet, the network is feed-forward and there is no incorporation of past information integrated over clicks. We also test VGG-Attention with human clicks (\textbf{VGG-Attention-humanclick}) and randomly generated click locations with strong spatial priors (\textbf{VGG-Attention-RandPrior}).


We considered several competitive baselines and existing methods of modeling temporal dynamics.
\vspace{-8mm}

  \paragraph{Human-fixations.} We were concerned about the variable viewing conditions in the MTurk experiments. Therefore, we conducted in-lab psychophysics measurements as a benchmark. In the in-lab experiment, after 500 ms fixation, a bounding box with a fixation cross in the center presented for 1,000 ms indicated the target position and cued subjects to attend to the hidden target location. To ensure that in-lab subjects paid attention to the hidden target location, we recorded their eye movements using an EyeLink D1000 system (SR Research, Canada). The image with the black box was shown for 200, 400, 800, or 1600 ms. Subjects freely moved their eyes; after stimulus offset, subjects said a single noun describing what the hidden target was. We recruited $4$ naive subjects (22 to 24 years old, 2 female), each one participating in 573 trials.
  \vspace{-4mm}

  \paragraph{SVM-category.} To study the effect of object co-occurrences, we used a binary vector of size $1\times C$ as input to a classifier, where the $i$th entry is 1 if there was an object from category $i$th in the image and 0 otherwise. We assume that the model had perfect information about \emph{all} the object labels (all objects were visible except for the hidden target). A multi-class support vector machine (SVM) classifier was used to predict the hidden target based on this vector.
  \vspace{-4mm}

  \paragraph{SVM-category-instances.} Extending SVM-category, we constructed a vector of size $1\times C$ where the $i$th entry contained the number $n$ of instances of the $i$th category in the image. A multi-class SVM classifier was used to predict the hidden target based on this vector.
  \vspace{-4mm}

  \paragraph{Hidden Markov Model (HMM).} To study the temporal dynamics over multiple clicks, we considered a Hidden Markov Model where we used all the training images in MSCOCO dataset to calculate the co-occurence matrix of size $C\times C$ as the transition probability matrix. We use normalized uniform vector of size $1\times C$ as the initial probability. We fine-tuned VGG16 on the MSCOCO dataset and used it for classifying the cropped region at the human clicked locations where the classification vector contributes to emission matrix. The Viterbi decoding algorithm \cite{blunsom2004hidden} was used for making inferences about the hidden target.
  \vspace{-4mm}

  \paragraph{DeepLab-Conditional Random Field (CRF)} One interesting solution to reason about the hidden target is to run state-of-the-art semantic segmentation algorithms and use majority voting on the predicted  labels over all pixels in the bounding box. We used the instantiation in DeepLab-CRF \cite{chen2017deeplab}.



\begin{figure*}[t]
 \centering
     \subfloat[Confusion matrix from ClickNet]{\includegraphics[width= 5.5cm, height = 4.0cm]{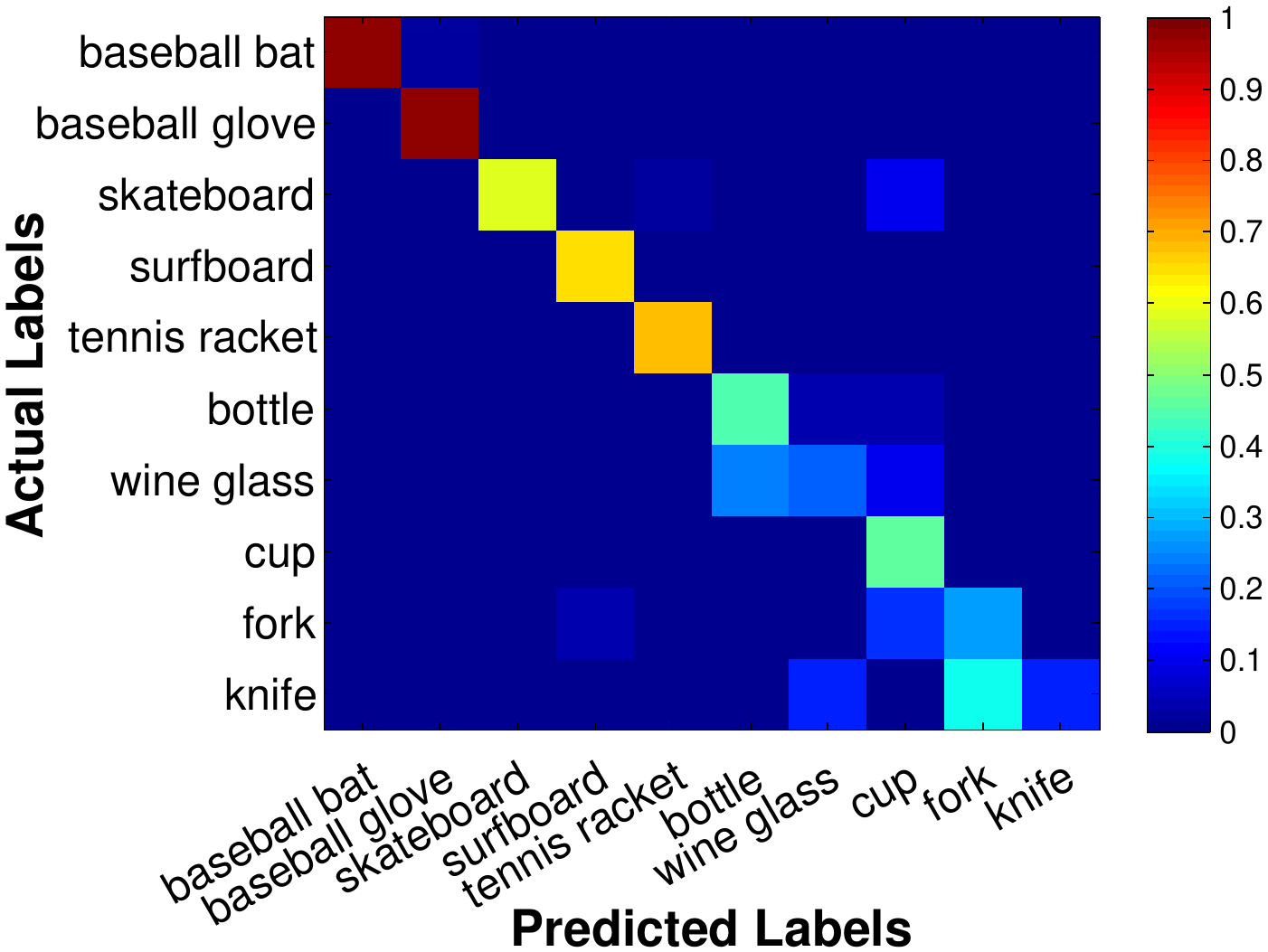}\label{fig:confuse}}\hspace{0.2cm}
     \subfloat[Accuracy versus context-object ratio]{\includegraphics[width= 8.0cm, height = 4.0cm]{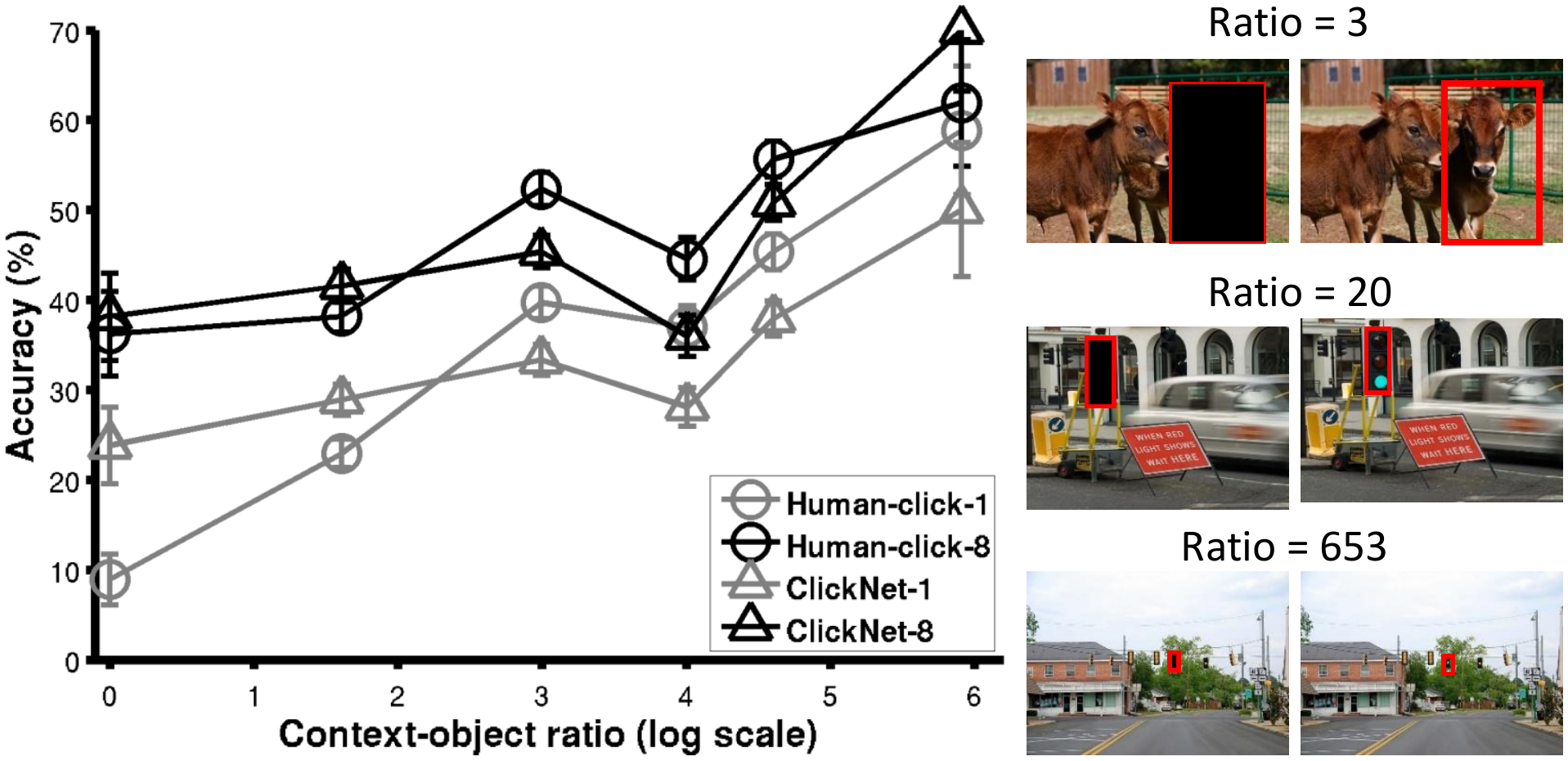}\label{fig:whatcoratio}}
     \caption{\textbf{Improvement in contextual reasoning accuracy with context-object ratio and patterns of mistakes}. (\textbf{a}) Partial view for illustrative purposes of confusion matrix showing the mistakes made by ClickNet among 10 of the 80 object categories in MSCOCO (Fig~\ref{fig:confusefull} in Appendix shows the complete confusion matrix with all 80 categories). The element in row $i$, column $j$ denotes the probability that ClickNet predicted label $j$ while the ground truth label was $i$ (see scale bar at right). The sum of probabilities in a row in the full confusion matrix (but not here) equals 1. (\textbf{b}) Human (circle) and model (triangle) accuracy for 1 click (gray) and 8 clicks (black) as a function of context-object ratio, shown in logarithmic scale (Sec.~\ref{subsec:evaluationmetrics}). Right: 3 example images with different context-object ratios. Only the images on the first column were shown (the second column is shown here for illustrative purposes only). Error bars indicate SEM across images. }
 \vspace{-4mm}\label{fig:confusionmatrixwhat}
 \end{figure*}

\vspace{-4mm}
\section{Results}\label{sec:results}

\vspace{-2mm}
\subsection{What: Importance sampling via attention and prior information}
\vspace{-2mm}
Subjects inferred the identity of the hidden target object with $36.7\pm0.9\%$ accuracy after 1 click (Fig~\ref{fig:overallperformance}, gray horizontal line). Performance showed a small, but significant improvement when allowing subjects 8 clicks, reaching  $48.4\pm0.9\%$ (Fig~\ref{fig:overallperformance}, black horizontal line, $p<10^{-9}$, two-tailed t-test, t=-6, df=2593). In-lab experiments corroborated these results showing accuracies of $48.1 \pm 0.9\%$ after 200 ms exposure and $56.3 \pm 0.9\%$ after 1,600 ms exposure to the images.

The same images that subjects saw were used to evaluate ClickNet (Figure~\ref{fig:overallperformance}). The ClickNet model showed a close approximation to human performance in the mturk experiments, reaching a top-1 classification performance of $33.3 \pm 0.9\%$ for 1 click and $45.0 \pm 0.9\%$ for 8 clicks. In both cases, performance was only slightly lower than human performance in 1 click ($p=0.17$, two-tailed t-test, t=1.4, df=1828) and 8 clicks ($p=0.17$, two-tailed t-test, t=1.4, df=1909). Performance for intermediate numbers of clicks is shown in (Figure~\ref{fig:overallperformancefull}). For all the computational models, random guessing would yield accuracy = $1.25\%$.

The worst performing model, VGG-Random, yielded performance above chance levels, emphasizing that even small amounts of high-resolution contextual data at arbitrary locations can help solve the problem. Yet, VGG-Random was well below ClickNet's performance ($p<10^{-5}$, two-tailed t-test, t=4, df=1144). Adding attention to the model (VGG-Attention) yielded only minimal improvement.
An important ingredient missing in VGG-Random and VGG-Attention is the informed location of the clicks. Humans and ClickNet do not sample the image randomly, but rather explore informative locations. Figure~\ref{fig:examples} shows qualitative examples of clicking patterns from humans and ClickNet. Both humans and ClickNet attend to salient regions on the images. For example, clicks often occur near traffic signs in the first example and near chairs and birds in the second example. Accordingly, substituting the random clicks for the human clicks into the VGG models (VGG-humanclick and VGG-Attention-humanclick) yielded a large performance boost. Conversely, substituting the ClickNet clicks with random clicks leads to large drop in performance when there is only 1 click, even when we artificially try to boost performance by constraining the clicks to be near the hidden target object (ClickNet-RandPrior). This effect is also evident with 2 clicks and 4 clicks (Fig.~\ref{fig:overallperformancefull}), but disappears with 8 clicks because there is already a lot of high resolution information in the image surrounding the hidden target, and ClickNet can integrate information over time to capitalize on it. Interestingly, the ClickNet sampling clicks are sufficiently close to human clicks that substituting the ClickNet clicks with human clicks does not improve performance (ClickNet-humanclick).

We considered several other comparative models (Fig.~\ref{fig:overallperformancefull}). Interestingly, using just a few clicks, ClickNet reaches performance that is essentially equivalent to that of VGG using a full resolution version of the entire image. Other comparative models (VGG-Blur, SVM-category, SVM-category-instances, HMM, DeepLab-CRF) showed above chance performance but their accuracies were well below ClickNet.

\vspace{-2mm}
\subsection{What: the more, the merrier}
\vspace{-2mm}

To investigate how much context information is needed to enhance recognition, we evaluated accuracy as a function of context-object ratio (Fig~\ref{fig:whatcoratio}, Sec~\ref{subsec:evaluationmetrics}). Images with higher context-object ratio contained more context information for inference, and yielded higher accuracy both for humans and models. Similarly, accuracy improved with increasing numbers of clicks (Fig~\ref{fig:overallperformance} and Fig~\ref{fig:whatcoratio}).


It is not just the quantity of context, but also the specific quality of contextual information that matters. In the real world, objects do not tend to appear in isolation but rather they co-vary with other objects. As ClickNet explores more regions on the image, it integrates information at previous clicked locations and learns associations of objects. The pattern of mistakes made by ClickNet is indicative of those associations (Fig~\ref{fig:confuse} and Fig~\ref{fig:confusefull}).  ClickNet often makes ``reasonable" wrong guesses when there is ambiguity in context reasoning, as humans do. For example, knife tends to be associated and therefore confused with spoon, fork, and wine glass, but knife seldom co-occurs with baseball bat or skateboard in these images.
\vspace{-2mm}

 \begin{figure*}[t]
 \centering
     \subfloat[Click-1 Consistency]{\includegraphics[width= 4.2cm, height = 4.1cm]{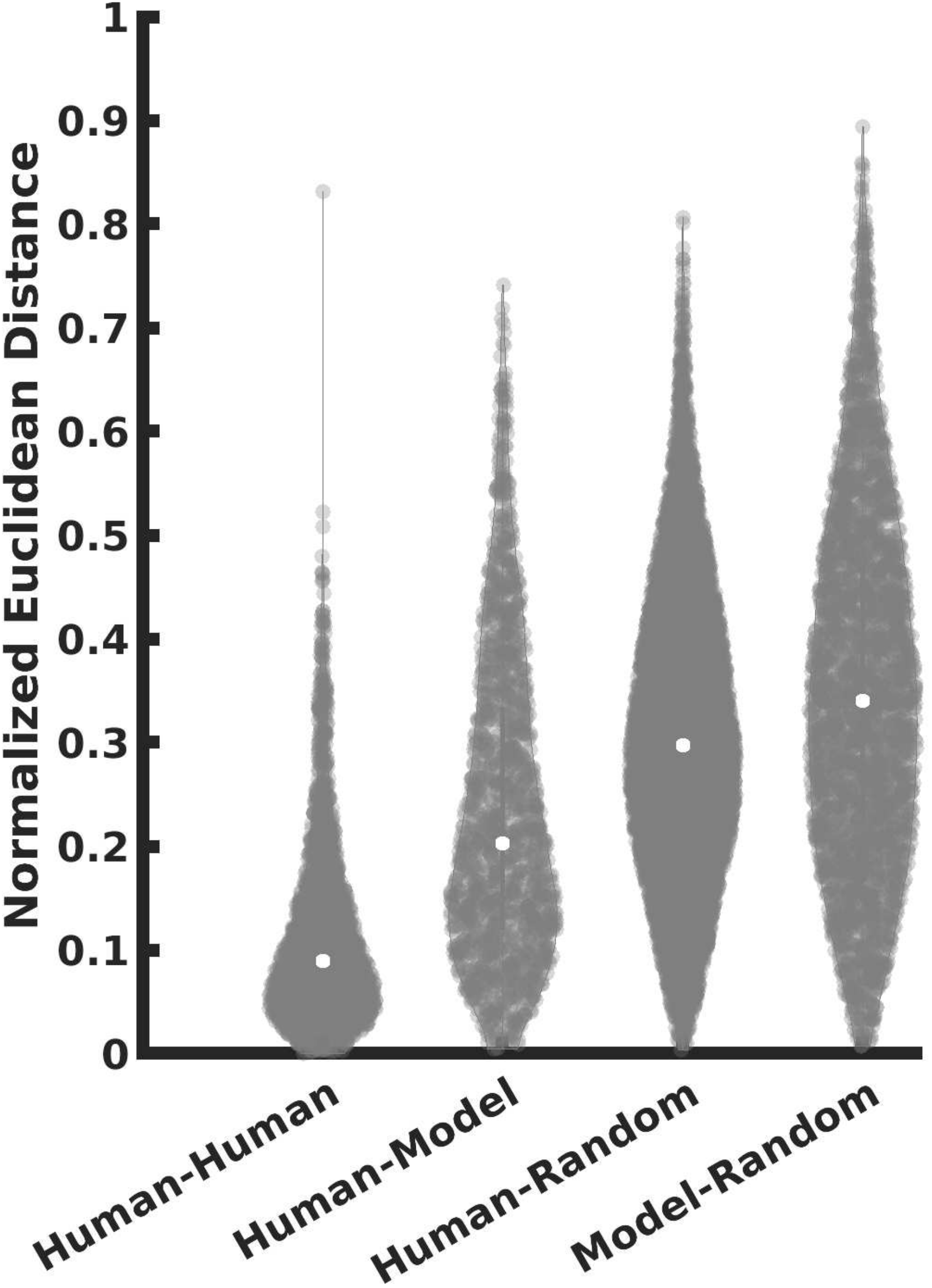}\label{fig:consistencyeuclid}}\hspace{0.3cm}
     \subfloat[Click-8 Consistency]{\includegraphics[width= 4.2cm, height = 4.1cm]{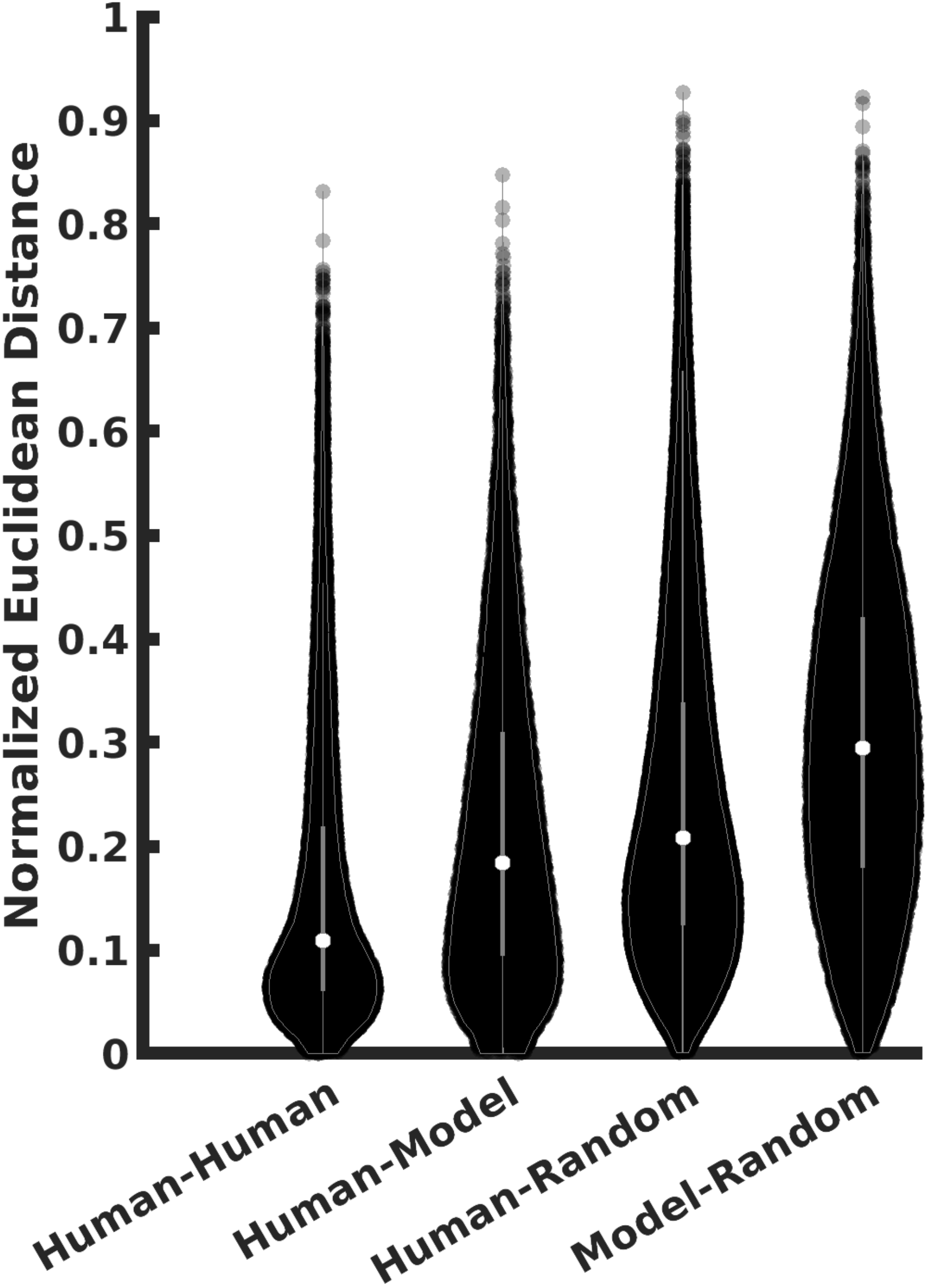}\label{fig:consistencyeuclid}}\hspace{0.3cm}
     \subfloat[Spatial distribution]{\includegraphics[width= 4.2cm, height = 4.1cm]{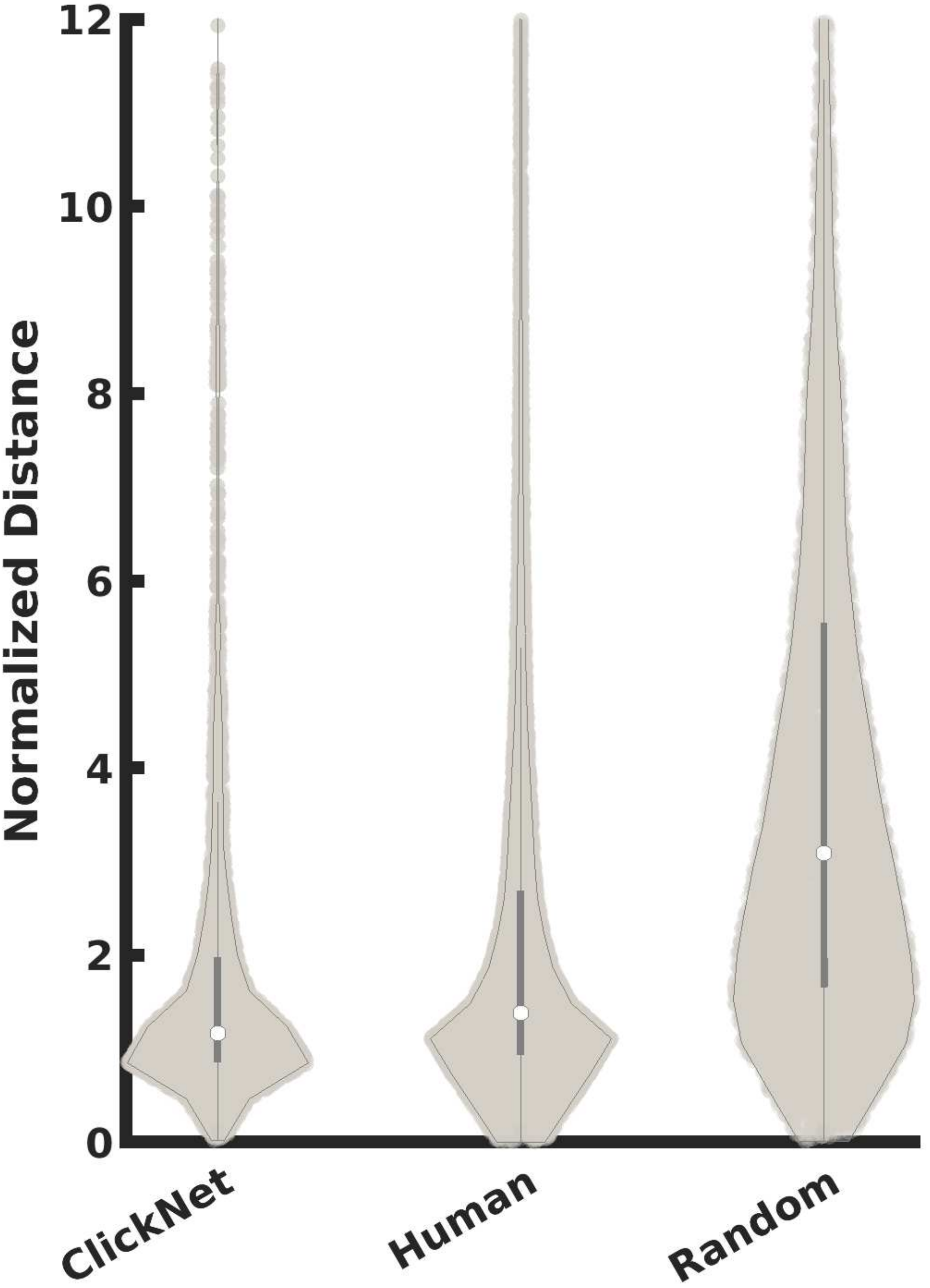}\label{fig:spatialprior}}
     \caption{\textbf{Model click locations were similar to human sampling.}
     (\textbf{a-b}) Consistency of click patterns between human subjects (human-human), consistency between humans and model (human-model), consistency between humans and random (human-random) and consistency between models and random (model-random) for 1 click (\textbf{a}) and 8 clicks (\textbf{b}) measured by the distribution of normalized Euclidean distances with respect to the diagonal of the image
     between any pairs of clicks by humans and ClickNet or random clicks. In each trial, we permute the sequence of mouse clicks between pairs of human and model clicks such that their sum of Euclidean distance is minimized across all clicks. The white circles denote the median of the distribution and the light grey bar denote the 1st and 3rd quartiles. (\textbf{b}) Euclidean Distance between click locations and center of the hidden target bounding box normalized by the diagonal of the hidden target bounding box.}
 \vspace{-4mm}\label{fig:consistencywhere}
 \end{figure*}

\vspace{-2mm}
\subsection{Where: consistency of human and model clicks}
\vspace{-2mm}
We hinted at the presumed similarity in the clicking patterns between humans and ClickNet based on the accuracy of the ClickNet-humanclick and ClickNet-RandPrior comparative models in Fig~\ref{fig:overallperformance}. To more directly assess whether ClickNet learned to sample the image to gather information about areas of contextual relevance, we directly quantified the similarity in clicking patterns (Figure~\ref{fig:consistencywhere}). To interpret the distances between human clicks and model clicks, we computed the degree of human-human consistency in the clicking patterns. The clicking patterns of ClickNet were overall similar to those made by humans. The model clicks were still different from the consistency between two humans (for 8 clicks, $p<10^{-15}$, two-tailed t-test, t=-30.4, df=29542); yet, the model clicks were much more similar to human clicks than random clicks (for 8 clicks, $p<10^{-15}$, two-tailed t-test, t=-50.3, df=35310).

\vspace{-2mm}
\subsection{Where: tendency of clicking nearby the target}\label{subsec:spatialprior}
\vspace{-2mm}
 There was a strong spatial bias towards clicking near the target for both humans and ClickNet (see examples in Figure~\ref{fig:examples}). To quantify this spatial bias, we computed the Euclidean distance between the clicked locations and the center of the bounding box, normalized by the diagonal of the bounding box (Figure~\ref{fig:spatialprior}). Humans tended to click within approximately one diagonal distance of the target box. Interestingly, although ClickNet does not take any human supervisory signal during training, ClickNet still learned to capture the tendency of clicking near the target.



We asked whether this spatial bias in sampling behavior is sufficient to explain performance in this task in a modified version of ClickNet. We removed the clicks dictated by the attention module and instead forced the clicks to be randomly distributed while still respecting the spatial distribution in Fig~\ref{fig:spatialprior} (ClickNet-RandPrior). Both ClickNet and ClickNet-humanclick surpassed ClickNet-RandPrior by 17\% and 10\% respectively (Figure~\ref{fig:overallperformance}). Similar results were obtained when using only the VGG architecture: VGG-Attention-humanclick was 16\% better than VGG-Attention-RandPrior (Fig~\ref{fig:overallperformancefull} in Appendix). Therefore, the spatial bias in clicking behavior is not sufficient to explain performance in this task. Sampling for context reasoning involves more than clicking near the target.
\vspace{-2mm}
\subsection{When: role of recurrent connections}
\vspace{-2mm}

Several lines of evidence support the importance of the recurrent network in the LSTM module in ClickNet. ClickNet outperformed the competitive baselines and state-of-the-art comparative methods to make inferences (Figure~\ref{fig:overallperformance} and Fig~\ref{fig:overallperformancefull} in Appendix). We tested whether the co-occurrence of object categories or the number of objects per category present in the image would be sufficient for context reasoning (SVM-category and SVM-category-instances). Even though these alternative models were exposed to full contextual information on the image and assumed perfect labeling of all objects in the image (except for the hidden target object), there was still a large overall performance drop in performance with respect to ClickNet. Moreover, graphical models for inference, such as Hidden Markov Model and DeepLab with Conditional Random Field (Sec~\ref{subsec:varaitionsclicknets}) failed to reach ClickNet's accuracy in this task (Fig~\ref{fig:overallperformancefull}).
The ablation studies eliminating the LSTM module further support the role of integrating information over multiple clicks in this task, as evidenced by the observation that ClickNet outperformed the VGG-Attention model.

\vspace{-2mm}
\section{Discussion}\label{sec:discussion}
\vspace{-2mm}
Here we quantitatively studied the role of contextual information in visual recognition in human observers and computational models in a task that involved inferring the identity of a hidden target object. Contextual influenced  recognition based on the amount of context, the specific location of contextual cues, and the dynamic sampling among salient visual features. We introduced a recurrent neural network model that combines a feed-forward visual stream module that extracts image features in a dynamic fashion, combined with an attention module to prioritize different image locations and select the next sampling step, and a recurrent LSTM module that integrates information over time and produces a label for the hidden object. Surprisingly, even though the model lacks the expertise that humans have in interacting with objects in their context, the model adequately predicts human sampling behavior and reaches almost human-level performance in this contextual reasoning task. The model opens the doors to examine more complex form of reasoning about scenes and how to integrate sequential sampling with prior knowledge.


\clearpage

\bibliographystyle{abbrv}
\bibliography{mengmibib}

\clearpage
\appendix
\section{Appendix}

\renewcommand{\thefigure}{S\arabic{figure}}
\renewcommand{\thetable}{S\arabic{table}}
\setcounter{figure}{0}

We provide supplementary figures and materials here. All labels in supplementary figures and tables are pre-fixed with letter S in front.

 \begin{figure*}[b]
 \begin{center}
 \includegraphics[width=13.5cm]{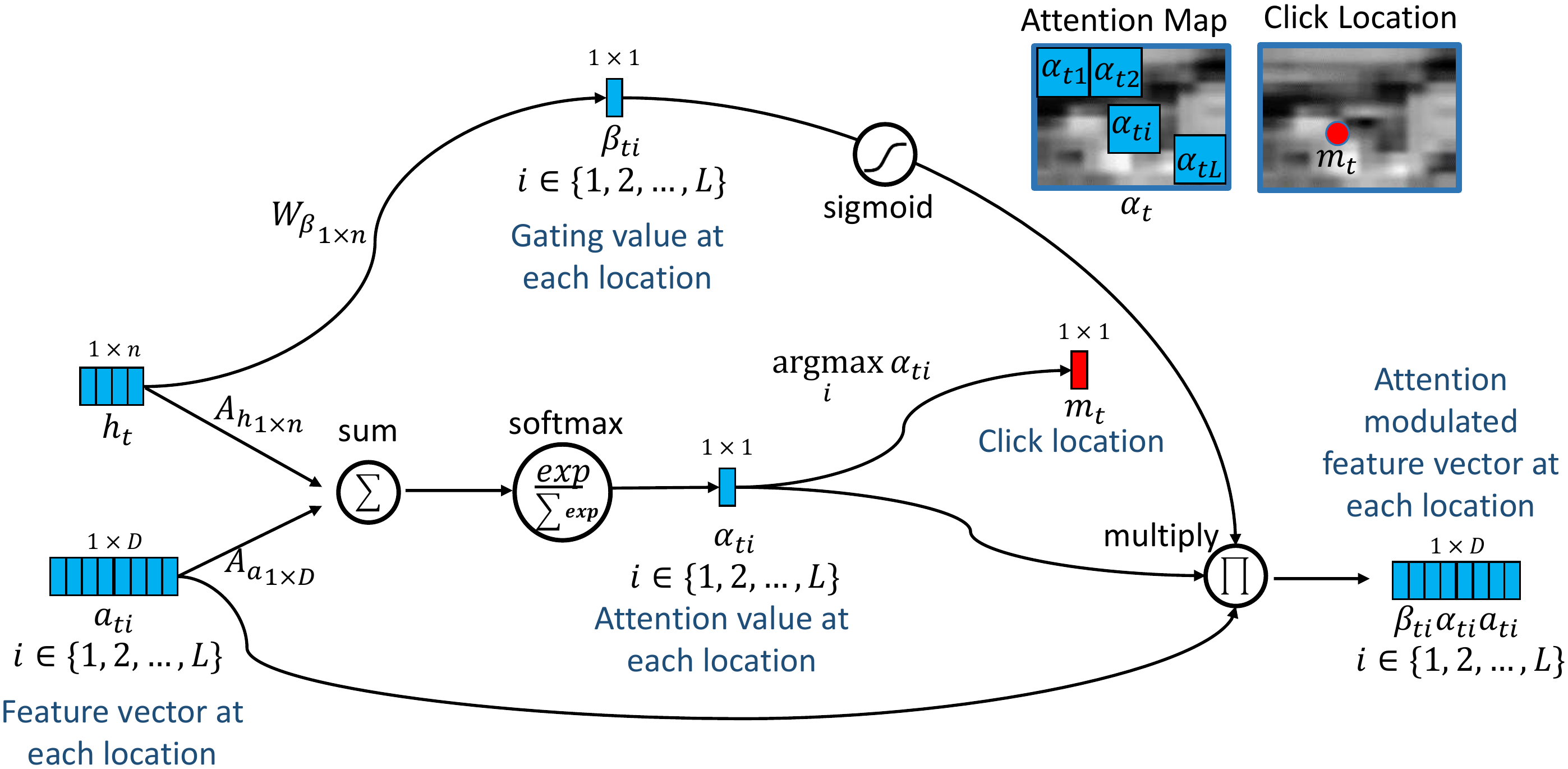}
 \end{center}
    \caption{\textbf{Schematic illustration of the attention module implementation}. Expanding on the overall ClickNet architecture shown in Fig~\ref{fig:model}, here we zoom into the attention module. The attention module takes as inputs the features at each location $a_{ti}$ and the output of the LSTM module $h_t$ and selects the next click location $m_t$ and a map that modulates the features at each location (see Section~\ref{sec:model} for a description of all the variables).
    }
 \vspace{-3mm}
 \label{fig:modelattention}
 \end{figure*}

 \begin{figure*}[b]
 \begin{center}
 \includegraphics[width=12.5cm]{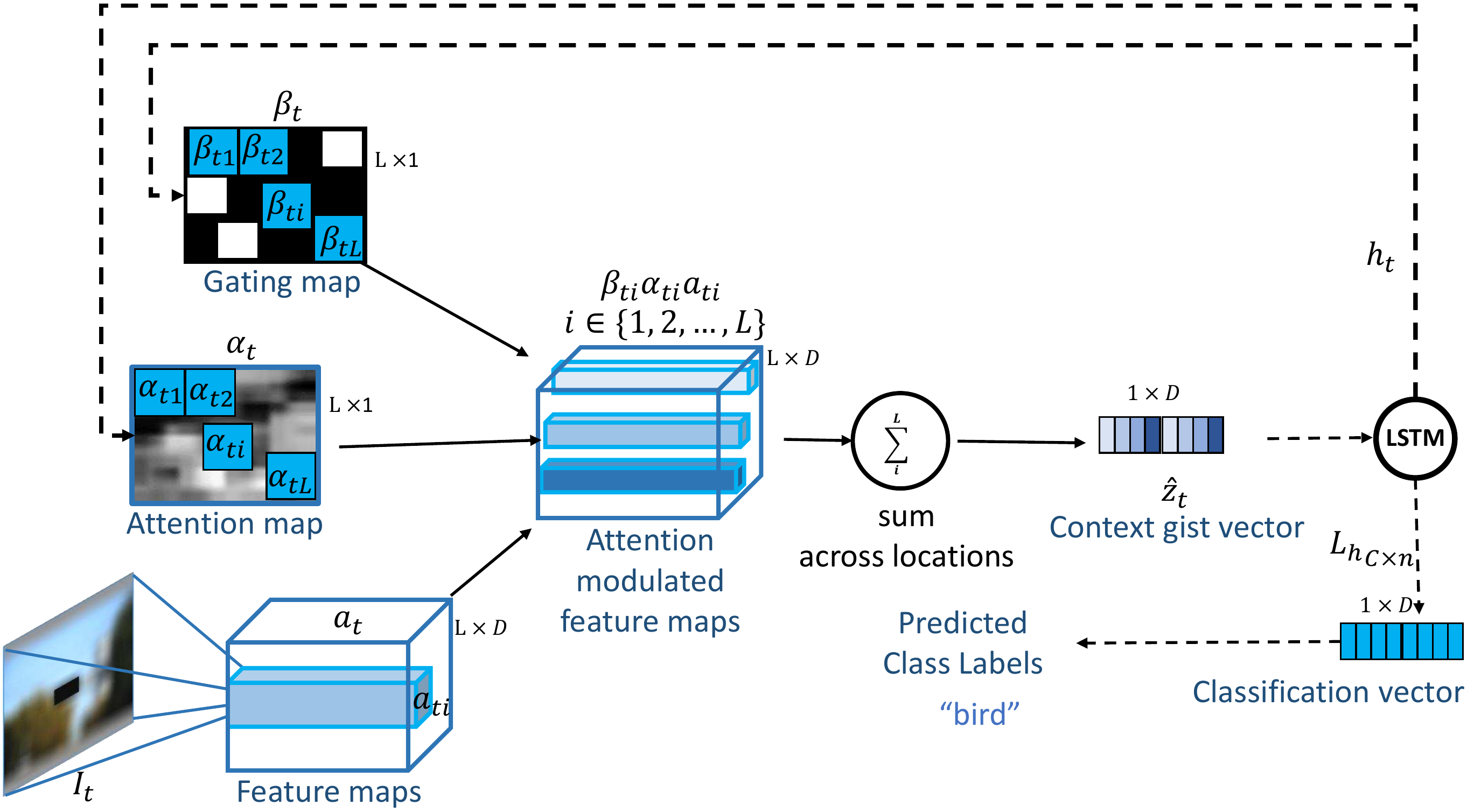}
 \end{center}
    \caption{\textbf{Schematic illustration of the LSTM module implementation}. Expanding on the overall ClickNet archietcture shown in Fig~\ref{fig:model}, here we zoom into the LSTM module. The LSTM module takes as input context gist vector $\mathbf{\widehat{z}_t}$ and integrates the information with the previous state to inform the attention module in the next time step via $h_t$ and to predict a class label (see Section~\ref{sec:model} for a description of all the variables).}
 \vspace{-3mm}
 \label{fig:modellstm}
 \end{figure*}

 \begin{figure*}[t]
 \begin{center}
 \includegraphics[width=12.5cm]{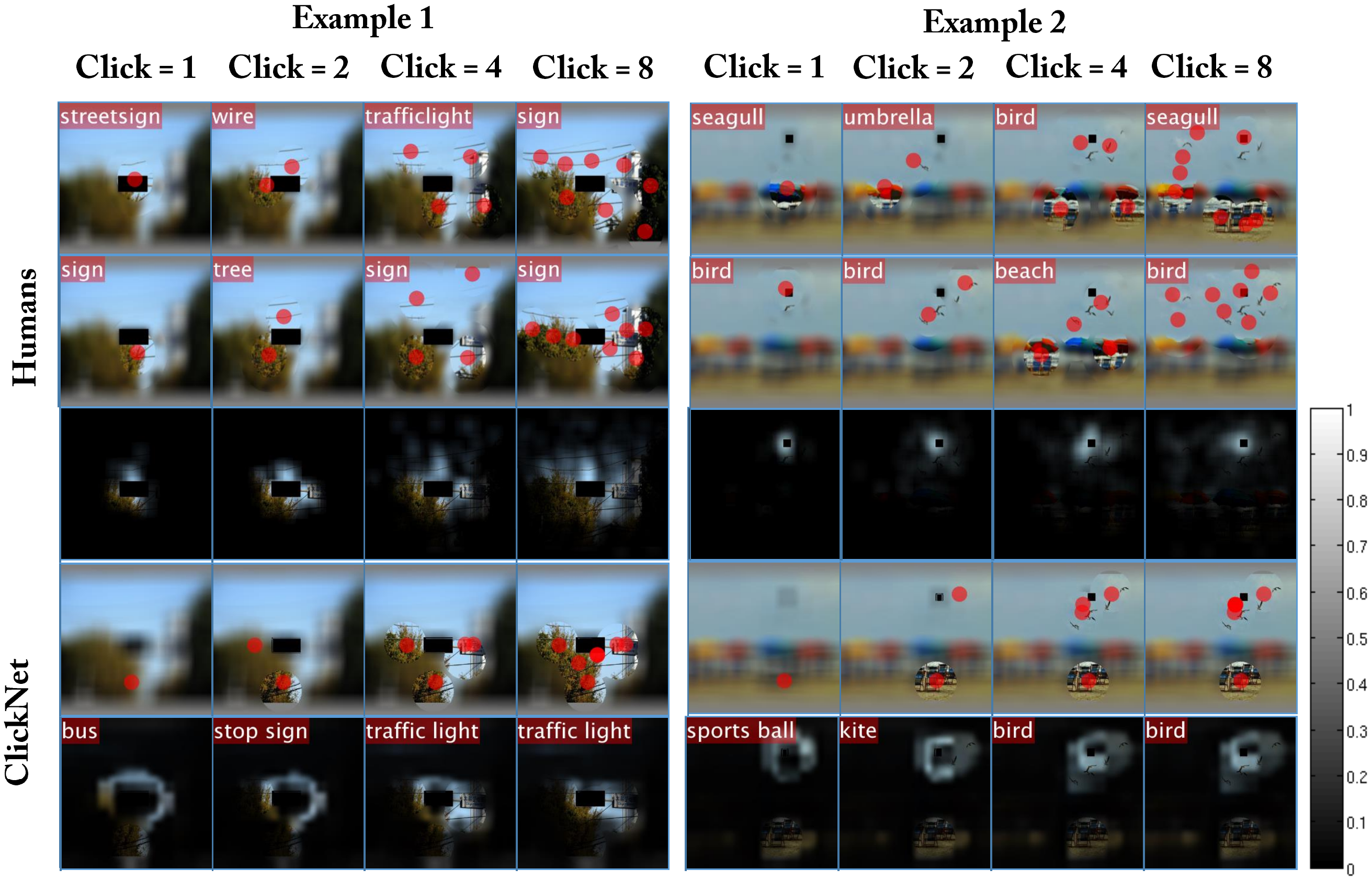}
 \end{center}
    \caption{\textbf{Example visualziation for humans and ClickNet.} This figure shows click locations and attention maps using the same format as Fig~\ref{fig:examples}, here adding results for 2 clicks and 4 clicks.}
 \vspace{-3mm}
 \label{Sfig:examplesall}
 \end{figure*}

 \begin{figure*}[t]
 \centering
     \subfloat[Click-1]{\includegraphics[width= 13cm]{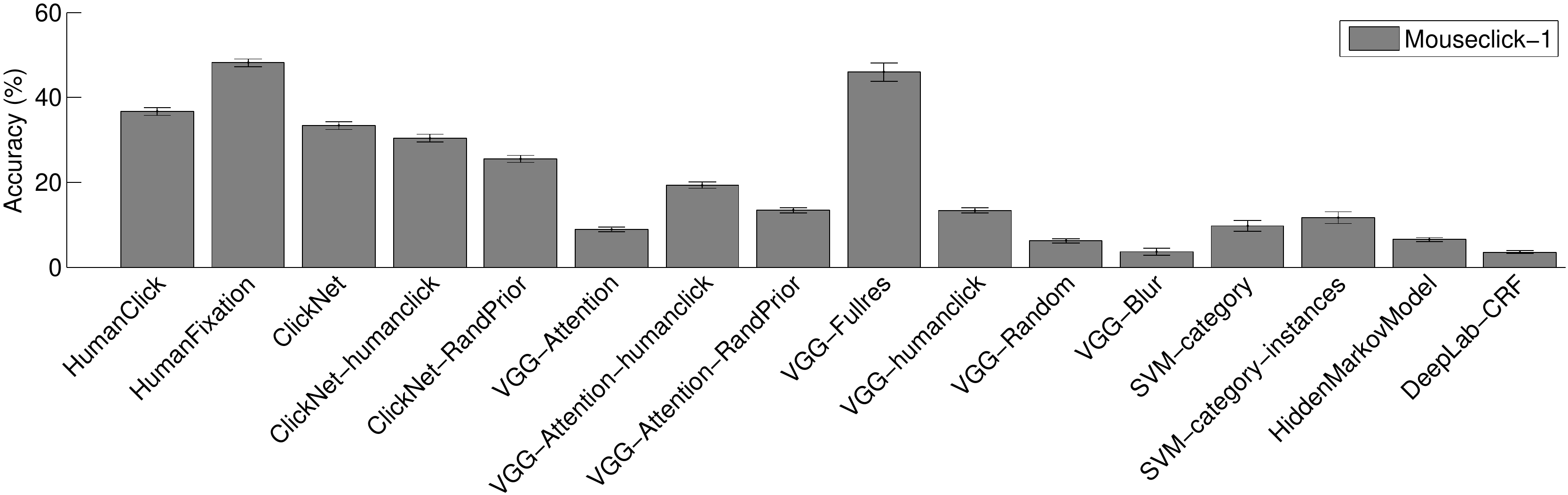}\label{fig:accuracyall1}} \hspace{0.1cm}
     \subfloat[Click-2]{\includegraphics[width= 13cm]{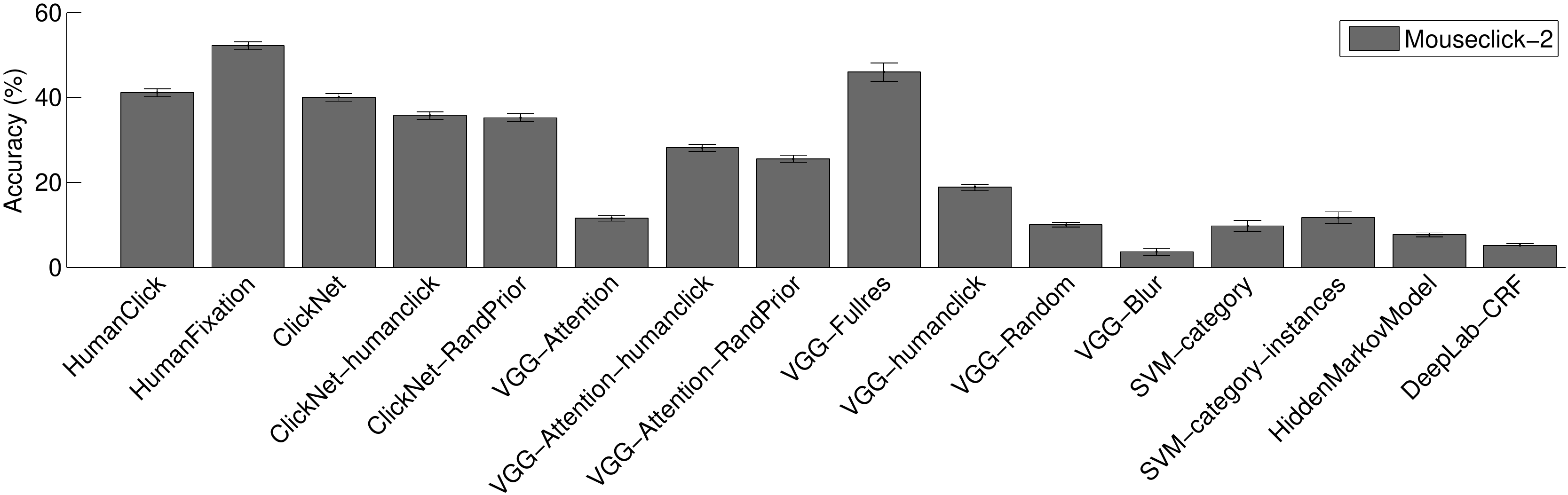}\label{fig:accuracyall2}} \hspace{0.1cm}
     \subfloat[Click-4]{\includegraphics[width= 13cm]{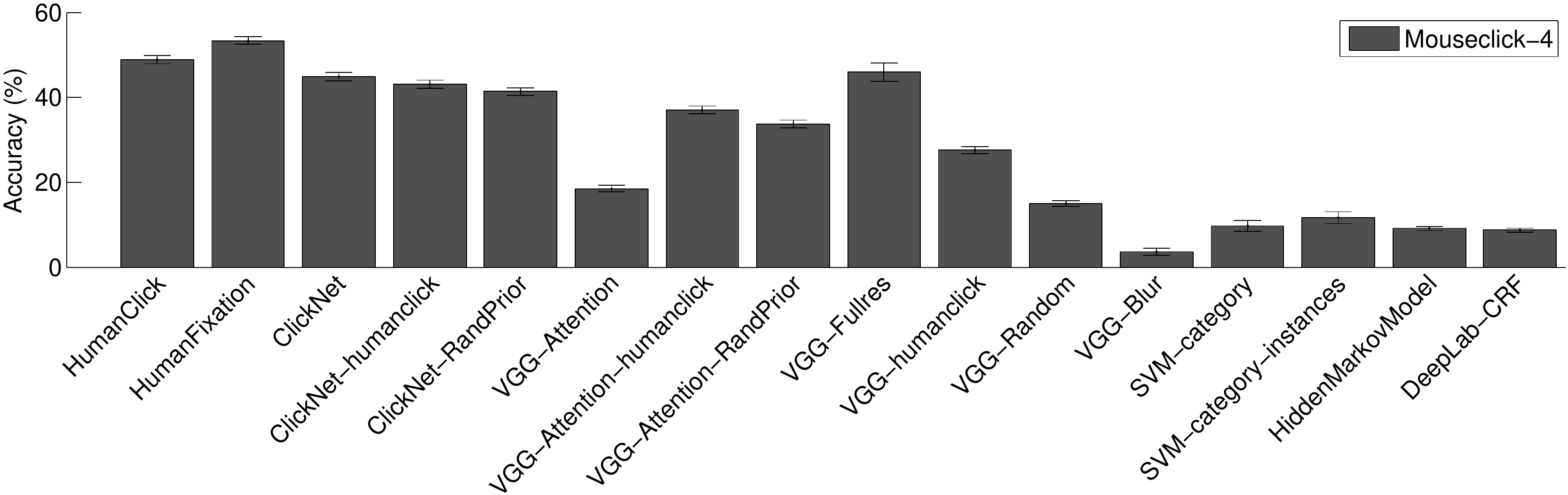}\label{fig:accuracyall3}} \hspace{0.1cm}
     \subfloat[Click-8]{\includegraphics[width= 13cm]{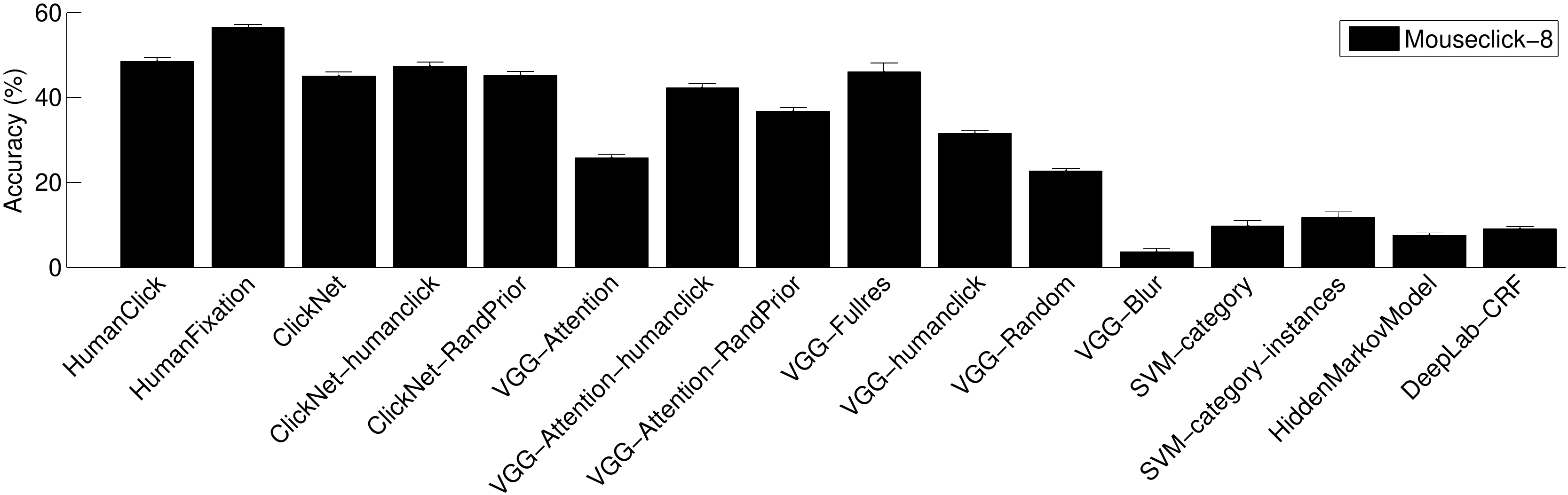}\label{fig:accuracyall4}}
     \caption{\textbf{Contextual reasoning accuracy of humans and models}. Expanding on the results in Fig.~\ref{fig:overallaccA}-b, here we add the results for 2 clicks and 4 clicks, as well as additional comparative models (Section~\ref{subsec:varaitionsclicknets}) describe each model).}
 \vspace{-4mm}\label{fig:overallperformancefull}
 \end{figure*}

\begin{figure*}[b]
 \begin{center}
 \includegraphics[width=13.5cm]{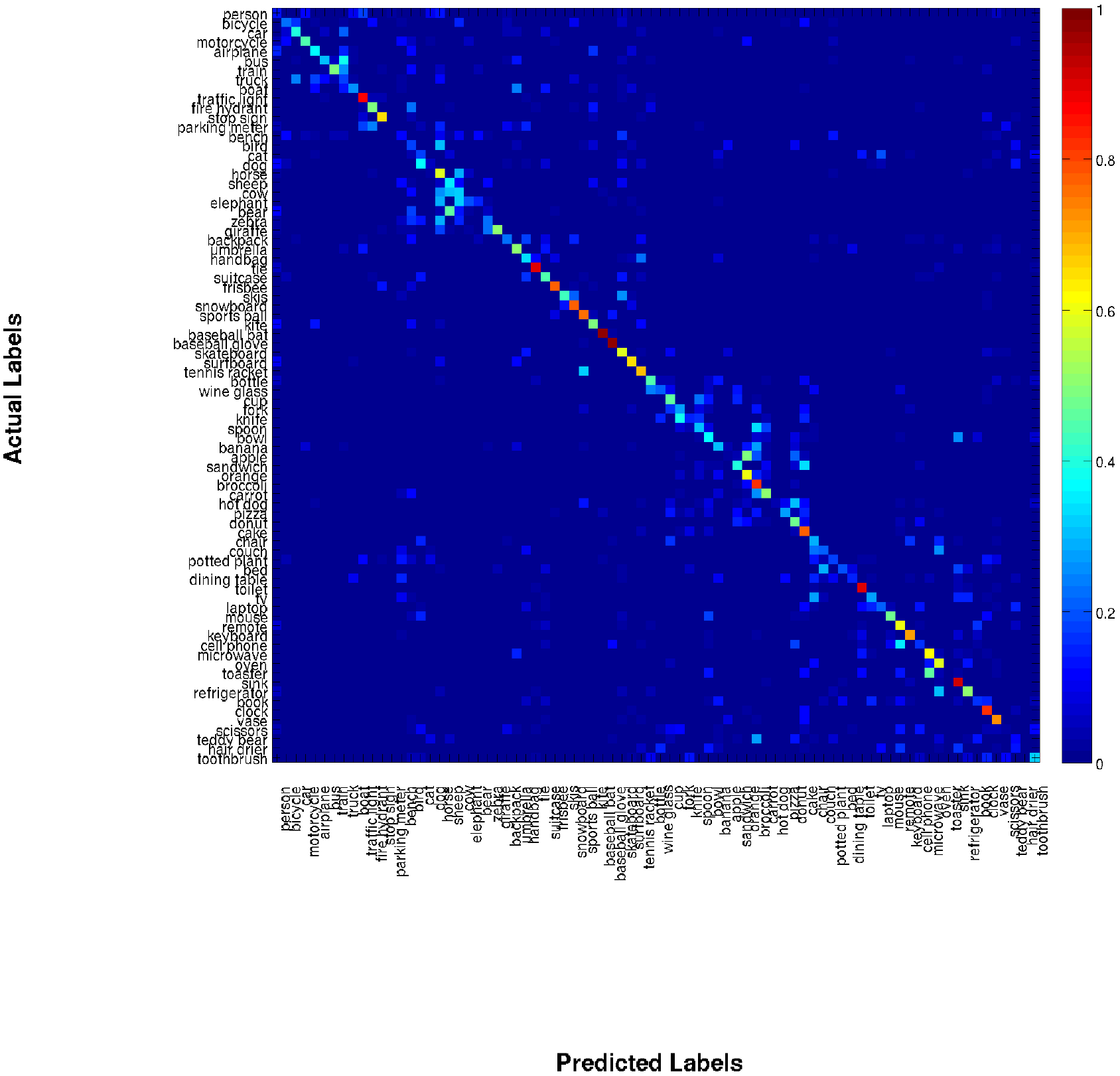}
 \end{center}
    \caption{\textbf{Confusion matrix for ClickNet with all click conditions}.   The format is the same as in Figure~\ref{fig:confuse}, except showing all 80 categories here. The element in row $i$, column $j$ denotes the probability that ClickNet predicted label $j$ while the ground truth label was $i$ (see scale bar on right). The sum of all probabilities in a row equals 1.}
 \vspace{-3mm}
 \label{fig:confusefull}
 \end{figure*}


\end{document}